\documentclass{article}

\usepackage{arxiv}

\usepackage[utf8]{inputenc} 
\usepackage[T1]{fontenc}    
\usepackage{hyperref}       
\usepackage{url}            
\usepackage{booktabs}       
\usepackage{amsfonts}       
\usepackage{nicefrac}       
\usepackage{microtype}      
\usepackage{lipsum}
\usepackage{graphicx}
\graphicspath{ {./images/} }
\usepackage{natbib}
\usepackage{amsmath}
\usepackage{multirow}
\usepackage{multicol}
\usepackage{rotating}
\usepackage{caption}
\usepackage{graphicx}
\usepackage{subcaption}

\DeclareUnicodeCharacter{0307}{\u{}}

\title{Virtual Nodes Improve Long-term Traffic Prediction}

\author{
 Xiaoyang Cao \\
  Department of Civil Engineering\\
  Tsinghua University\\
  Beijing 100084, China \\
  \texttt{cao-xy21@mails.tsinghua.edu.cn} \\
   \And
 Dingyi Zhuang \\
  Department of Civil and Environmental Engineering\\
  Massachusetts Institute of Technology\\
  Cambridge, MA 20139 \\
  \texttt{dingyi@mit.edu} \\
  \And
 Jinhua Zhao \\
  Department of Urban Studies and Planning\\
  Massachusetts Institute of Technology\\
  Cambridge, MA 20139 \\
  \texttt{jinhua@mit.edu} \\
  \And
 Shenhao Wang \\
  Department of Urban and Regional Planning\\
  University of Florida\\
  Gainesville, FL 32611 \\
  \texttt{shenhaowang@ufl.edu} \\
}

\begin{document}
\maketitle
\begin{abstract}
Effective traffic prediction is a cornerstone of intelligent transportation systems, enabling precise forecasts of traffic flow, speed, and congestion.
While traditional spatio-temporal graph neural networks (ST-GNNs) have achieved notable success in short-term traffic forecasting, their performance in long-term predictions remains limited. This challenge arises from over-squashing problem, where bottlenecks and limited receptive fields restrict information flow and hinder the modeling of global dependencies.
To address these challenges, this study introduces a novel framework that incorporates virtual nodes—additional nodes added to the graph and connected to existing nodes—to aggregate information across the entire graph within a single GNN layer. 
Our proposed model incorporates virtual nodes by constructing a semi-adaptive adjacency matrix. This matrix integrates distance-based and adaptive adjacency matrices, allowing the model to leverage geographical information while also learning task-specific features from data.
Experimental results demonstrate that the inclusion of virtual nodes significantly enhances long-term prediction accuracy while also improving layer-wise sensitivity to mitigate the over-squashing problem.
Virtual nodes also offer enhanced explainability by focusing on key intersections and high-traffic areas, as shown by the visualization of their adjacency matrix weights on road network heat maps.
Our advanced approach enhances the understanding and management of urban traffic systems, making it particularly well-suited for real-world applications.
\end{abstract}


\section{Introduction}
Forecasting traffic conditions, such as flow, speed, and congestion, plays a central role in intelligent transportation systems (ITS).
This capability is essential for enhancing urban traffic efficiency and mitigating congestion, which are critical challenges in modern cities.
The task relies on analyzing extensive datasets collected from road sensors, GPS devices, and historical traffic records to identify and model traffic patterns. 
Advanced methodologies, including statistical models, machine learning techniques, and deep learning frameworks, have significantly contributed to progress in this area.

Graph Neural Networks (GNNs) have demonstrated remarkable success in traffic prediction, particularly in short-term forecasting. However, their ability to handle long-term predictions remains constrained.
According to \citet{wang2020long}, forecasts with a time span of less than one hour are classified as short-term, while those extending beyond one hour are regarded as medium or long-term \citep{hou2014traffic, yu2017spatio}.
Long-term traffic prediction requires not only capturing long-range correlations but also effectively incorporating global spatial and temporal information.
The message-passing mechanism of Graph Neural Networks (GNNs), which operates by recursively aggregating information from neighboring nodes, presents a significant limitation as it primarily focuses on local connections, often neglecting global relationships.
The recursive structure of GNNs gives rise to the \textit{over-squashing} problem \citep{alon2020bottleneck}, in which information from distant nodes becomes compressed into fixed-size representations as it propagates through the network layers.
Various research has extensively documented the limitations of fixed-size vector representations, showing that they create information bottlenecks that hinder the effective capture of long-range dependencies. 
Moreover, as the distance between nodes increases, stacking additional GNN layers often leads to performance degradation, making it increasingly difficult to retain detailed information from distant nodes \citep{alon2020bottleneck, banerjee2022oversquashing, di2023over}.
Given the critical role of global and spatially long-range dependencies—where distant locations can substantially influence future traffic conditions—it is essential to incorporate global spatial information into long-term prediction models to improve their accuracy and performance.

Virtual nodes are an effective graph rewiring technique designed to enhance the representation of long-range correlations in graphs.
Unlike conventional graph rewiring, which dynamically modifies the graph’s structure to optimize information flow, virtual nodes introduce additional nodes without altering the original graph topology. 
These virtual nodes are connected to all existing nodes, with their graph signals and edge weights either manually specified or automatically learned based on task requirements. 
Acting as intermediaries, virtual nodes facilitate the efficient aggregation of global information, mitigating the over-squashing problem that arises when information from distant nodes is compressed and lost during propagation. 
By enabling the integration of information from the entire graph within a single GNN layer, virtual nodes streamline the message-passing process, allowing models to effectively capture long-range dependencies and critical spatial relationships.

Virtual nodes have proven effective in enhancing molecular graph representation for drug discovery \citep{li2017learning} and improving graph-level pattern recognition in classification tasks \citep{gilmer2017neural, pham2017graph, li2017learning, liu2022boosting}. 
Despite their demonstrated success in these domains, the application of virtual nodes in traffic prediction remains largely unexplored.
Virtual nodes are particularly well-suited for traffic prediction due to their ability to address the challenges of long-term forecasting. 
By efficiently capturing long-range dependencies across traffic networks, they enable models to predict traffic conditions over extended periods with greater accuracy. 
Moreover, virtual nodes provide an intuitive explainability to the model. They can be viewed as intelligent transportation hubs that gather information from different parts of the traffic network and broadcast it throughout the system. 
Functioning as a central processing unit, virtual nodes synthesize data to offer a holistic view of the entire traffic system through GNN's message-passing mechanism. 

In this study, we propose an advanced traffic prediction model that incorporates virtual nodes within a semi-adaptive adjacency matrix framework. Our model aims to address the limitations of traditional ST-GNNs, particularly their difficulty in capturing long-range spatial dependencies crucial for accurate long-term traffic prediction. The primary contributions of this study are as follows:

\begin{itemize} 
\item \textbf{Virtual Node Integration:} 
We introduce virtual nodes to address the \textit{over-squashing} problem in ST-GNNs, enhancing their ability to capture long-range spatial correlations. This method is versatile and can be seamlessly integrated into existing ST-GNN architectures. 
\item \textbf{Semi-Adaptive Adjacency Matrix:} 
A semi-adaptive adjacency matrix framework is proposed, combining distance-based and adaptive adjacency matrices. This hybrid design effectively learns the connection weights for virtual nodes, striking a balance between model complexity and adaptability to improve the integration of virtual nodes. 
\item \textbf{Enhanced Long-Term Prediction Accuracy and Sensitivity:} 
Experimental results demonstrate that incorporating virtual nodes significantly improves long-term traffic prediction accuracy. Moreover, the layer-wise sensitivity of ST-GNNs is notably increased, indicating stronger connections between distant nodes and effectively mitigating the \textit{over-squashing} problem. 
\item \textbf{Explainability and Visualization:} Our approach provides improved explainability by visualizing the real-to-virtual adjacency matrix and generating road network heat maps. These visualizations reveal spatial patterns in connection weights, with higher weights automatically assigned to traffic-critical regions, such as major intersections and junctions. 
\end{itemize}

The rest of the paper is organized as follow: Section \ref{sec:problem} explains GNNs, ST-GNNs, and the challenge of long-term traffic prediction; Section \ref{sec:literature} reviews relevant research on traffic prediction and graph rewiring techniques; Section \ref{sec:method} details our approach, including virtual node integration and the semi-adaptive adjacency matrix; Section \ref{sec:experiment} presents the dataset, ablation studies, model comparisons, sensitivity analysis, and visualization results; and Section \ref{sec:conclusion} summarizes findings, discusses implications, and suggests future research directions.

\section{Problem description and preliminaries}
\label{sec:problem}
\subsection{Graph neural networks}

\textit{Graph Neural Networks} (GNNs) \citep{scarselli2008graph} have recently emerged as one of the most prominent machine learning architectures for analyzing graph-structured data. They have been widely adopted in various domains where data is naturally represented as graphs, such as traffic prediction \citep{wu2021inductive, zhuang2022uncertainty, bai2020adaptive}, biochemistry \citep{gilmer2017neural, jumper2021highly}, and social network analysis \citep{fan2019graph, he2020lightgcn}.

GNNs recursively learn vector representations for each node by aggregating information from its neighbors. 
To formalize this process, we adopt the following notational conventions: calligraphic fonts denote sets, bold uppercase letters represent matrices, bold lowercase letters denote vectors, and subscripts are used for indexing nodes and time steps.

Let the graph be defined as $\mathcal{G} = (\mathcal{V}, \mathcal{E}, \mathbf{A})$, where $\mathcal{V}$ represents the set of nodes (locations), $\mathcal{E}$ denotes the set of edges, and $\mathbf{A} \in \mathbb{R}^{|\mathcal{V}| \times |\mathcal{V}|}$ is the adjacency matrix encoding spatial relationships between nodes.
The specific definition of the adjacency matrix $\mathbf{A}$ used in this study is given in Section \ref{subsec:dataset}. 
In each layer $i$, the node features are updated as follows \citep{scarselli2008graph}:
\begin{equation}
\mathbf{h}_v^{(i)} = \text{UPDATE}^{(i)} \left( \mathbf{h}_v^{(i-1)}, \text{AGG}^{(i)} \left( \left\{ \mathbf{h}_u^{(i-1)} \mid u \in \mathcal{N}(v) \right\} \right) \right),
\label{eq:message_passing}
\end{equation}
where \(\text{UPDATE}^{(i)}\) and \(\text{AGG}^{(i)}\) are differentiable parameterized functions, and \(\mathbf{h}_v^{(0)}\) represents the initial node feature. 
Here, \(\mathcal{N}(v)\) denotes the set of neighboring nodes of node \(v\). 
For node-level tasks such as traffic prediction, the output in the final layer \( N \) is used directly:
\begin{equation}
\hat{y}_v = \mathbf{W} \mathbf{h}_v^{(N)}.
\end{equation}

While GNNs are effective for modeling static relationships within a topological structure, they are inherently limited to processing data at fixed time points and cannot capture the patterns of feature evolution over time in sequential data, which are crucial for traffic prediction.
To overcome this limitation, Spatio-Temporal Graph Neural Networks (ST-GNNs) have been proposed. These models extend the capabilities of traditional GNNs by incorporating temporal dependencies, often leveraging architectures such as Long Short-Term Memory Networks (LSTMs) \citep{hochreiter1997long} and Transformers \citep{vaswani2017attention}.
This enhancement allows ST-GNNs to effectively model the spatio-temporal dynamics required for accurate traffic prediction.

\subsection{ST-GNNs for traffic prediction}

ST-GNNs have been widely applied in traffic prediction tasks \citep{bai2020adaptive, li2018diffusion, wu2021inductive, zhuang2022uncertainty, han2019predicting, han2021dynamic, Kong2020STGAT:Forecasting, yu2017spatio, shao2022decoupled},  demonstrating strong capabilities in modeling complex spatio-temporal dynamics.
ST-GNNs address the spatial and temporal components of traffic data through distinct mechanisms. The spatial component is handled by GNNs, which aggregate and extract features by capturing relationships between nodes (locations). Meanwhile, the temporal component leverages methods such as Gated Recurrent Units (GRUs) \citep{cho2014learning}, Temporal Convolutional Networks (TCNs) \citep{lea2017temporal}, and Transformers \citep{vaswani2017attention} to model sequential patterns over time.
Specifically, for each time step, GNNs are used to extract spatial features and generate graph embeddings, which are then fed into the temporal module. These temporal models take embeddings from a fixed number of past time steps as input and predict graph signals for specified future time steps. This integration of GNNs and temporal techniques allows ST-GNNs to effectively capture both spatial and temporal dependencies, ensuring accurate traffic predictions.

We denote the traffic dataset inputs as $\mathcal{X} \in \mathbb{R}^{|\mathcal{V}| \times t}$, where $t$ is the number of time steps. 
The objective is to predict the target value $\mathcal{Y}_{1:|\mathcal{V}|, t:t+k}$ for the future $k$ time steps, given all past data up to time $t$, denoted as $\mathcal{X}_{1:|\mathcal{V}|, 1:t}$. 
Given a graph $\mathcal{G} = (\mathcal{V}, \mathcal{E}, \mathbf{A})$, the goal of the ST-GNN models is to design a model $f_{\theta}$, parametrized by $\theta$, such that:
\begin{equation}
    \hat{\mathcal{Y}}_{1:|\mathcal{V}|, t:t+k} = f_{\theta}(\mathcal{X}_{1:|\mathcal{V}|, 1:t}; \mathcal{G}) = f_{\theta}(\mathcal{X}_{1:|\mathcal{V}|, 1:t}; \mathcal{V}, \mathcal{E}, \mathbf{A}),
    \label{eq:GNN}
\end{equation}
where $\hat{\mathcal{Y}}_{1:|\mathcal{V}|, t:t+k}$ is the predicted target value. 
The loss function like Mean Absolute Error (MAE) is used to optimize the model parameters $\theta$ during training.

\subsection{Long-term traffic prediction}

\begin{figure}[htbp]
    \centering
    \includegraphics[width=0.7\linewidth]{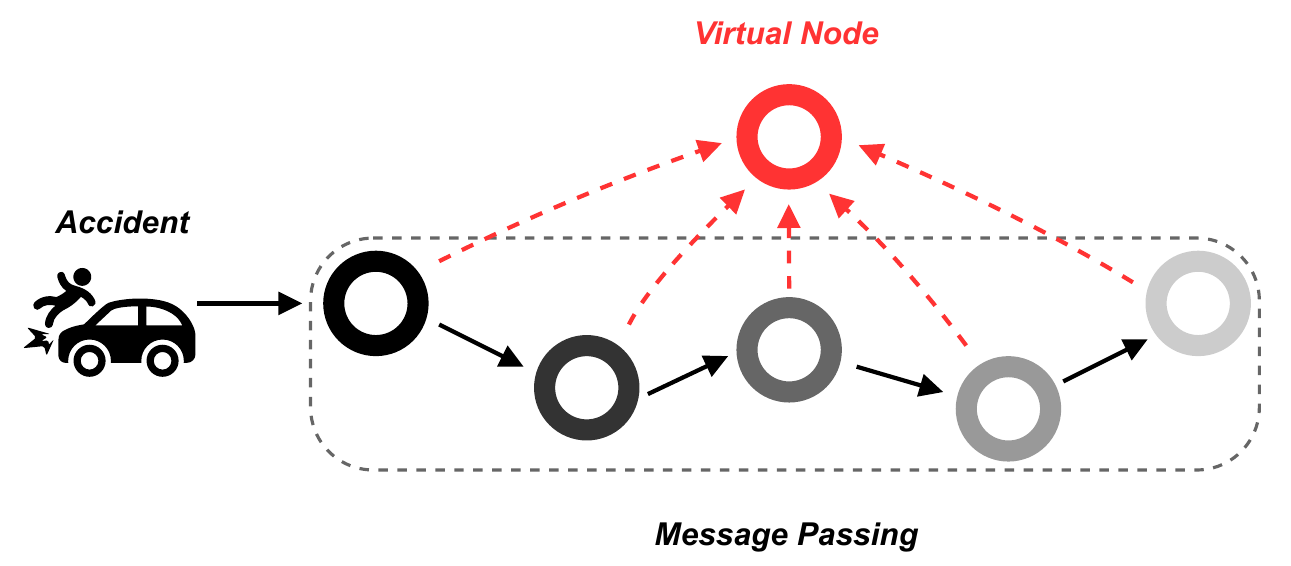}
    \caption{GNN with Virtual Node Augmentation}
    \label{fig:vn_aug}
\end{figure}

Previous research has primarily focused on using spatio-temporal graph neural networks (ST-GNNs) for short-term traffic prediction, typically within a 60-minute horizon when the temporal granularity is 5 minutes (\cite{zhuang2022uncertainty, han2021dynamic, Kong2020STGAT:Forecasting, yu2017spatio, shao2022decoupled}).
This limitation arises because the information propagation speed between nodes in ST-GNNs is relatively slow, making it challenging to accurately predict long-term traffic states. 
Due to the nature of the GNN message-passing mechanism, each layer aggregates information only from neighboring nodes. 
As shown in Equation (\ref{eq:message_passing}), each node \( v \) updates its representation \( \mathbf{h}_v^{(i)} \) by aggregating features from its immediate neighbors \( \mathcal{N}(v) \) at the previous layer \( i - 1 \). 
Information from nodes that are \( l \)-hops away requires at least \( l \) layers for their information to reach node \( v \). 
Consequently, capturing long-range dependencies requires stacking many layers, which can lead to the \textit{over-squashing} problem.

In the context of traffic prediction, the over-squashing problem becomes particularly evident. 
Consider an example in Figure \ref{fig:vn_aug}, where an accident occurs at the westernmost node of a road network. 
The impact gradually propagates throughout the traffic flow along the network. Traditional ST-GNNs, which aggregate information from neighboring nodes, require multiple layers to convey the impact from the westernmost to the easternmost node. 
This slow propagation results in the compression of information into fixed-size representations as it passes through the layers, causing critical details from distant nodes to be lost or diminished. 
This latency hinders the ability of ST-GNNs to promptly capture the effects of sudden incidents on traffic flow, thereby limiting their efficacy for long-term prediction.

This study introduces a method based on Virtual Nodes (\cite{gilmer2017neural}) to accelerate the learning of spatio-temporal relationships in ST-GNNs. 
Virtual Nodes are connected to all real nodes in the network, enabling them to aggregate information from across the entire graph in a single GNN layer. 
Weights between the Virtual Nodes and other nodes in adjacency matrix $\mathbf{A}$ are determined through an adaptive learning mechanism, ensuring appropriate weight allocation to diverse information types. 
The incorporation of Virtual Nodes accelerates the message passing process in ST-GNNs, improving the accuracy of long-term traffic prediction.

\section{Literature Review}
\label{sec:literature}
After introducing the problem and concept, and before we delve into our proposed methodology, we review the existing work on ST-GNN for traffic prediction, as well as the current state of technology regarding virtual nodes.

\subsection{Traffic prediction with ST-GNN}

With the advancement of deep learning in recent years, researchers have explored various frameworks for traffic prediction, including Convolutional Neural Networks (CNNs) (\cite{yao2018deep,Yao2019RevisitingPrediction}), Recurrent Neural Networks (RNNs) (\cite{bai2020adaptive, li2018diffusion}), and ST-GNNs (\cite{bai2020adaptive, li2018diffusion, wu2021inductive, zhuang2022uncertainty, han2019predicting, han2021dynamic, Kong2020STGAT:Forecasting, yu2017spatio, shao2022decoupled}). 
ST-GNNs, in particular, have garnered attention for their ability to simultaneously handle spatial and temporal dependencies. 
This capability is crucial for accurately modeling real-world traffic conditions, which constantly fluctuate across time and space, necessitating dynamic handling of spatio-temporal correlations. 
Spatially, ST-GNNs convert road network topologies into graph structures, allowing them to handle a wide array of non-Euclidean spatial data. 
Temporally, they effectively learn the non-linear and periodic patterns in traffic data, leading to robust performance in complex traffic prediction tasks.

DCRNN (\cite{li2018diffusion}) is one of the first models that simultaneously learns spatial and temporal dependencies, using GNNs to process spatial relationships and RNNs to process temporal dynamics.
STGCN (\cite{yu2017spatio}) utilizes CNN to concurrently manage spatial and temporal dependencies, offering higher computational efficiency compared to RNNs. 
AGCRN (\cite{bai2020adaptive}) and Graph WaveNet (\cite{wu2019graph}) introduce adaptive adjacency matrices, allowing for the dynamic adjustment of weights to capture the hidden spatial dependencies more accurately. 
DGCRN (\cite{li2023dynamic}) extracts dynamic characteristics from node attributes, enabling the generation of time-varying adjacency matrices. 
GMAN (\cite{zheng2020gman}) leverages a self-attention mechanism to effectively capture global spatio-temporal correlations, enhancing GNN's ability to process global information.

However, the architecture of ST-GNNs leads to the \textit{over-squashing} problem (\cite{alon2020bottleneck}), where information from distant nodes is compressed into fixed-size representations as it propagates through the layers. 
This compression can cause important details from distant nodes to be lost or diminished, resulting in a bottleneck that limits the model's ability to capture long-range dependencies effectively. 
Consequently, ST-GNNs face challenges in processing global information and long-term interactions, which significantly restricts their predictive capabilities. 
To address these limitations, researchers have proposed the Graph Transformers (GTs) to handle global information by applying global attention mechanisms (\cite{chen2022structure, kim2022pure, dwivedi2022long, wu2021representing}). 
However, all transformers suffer from their quadratic space and memory requirements, thus limiting their practical applicability in larger networks. 
Therefore, techniques have emerged that focus on modifying the network structure to effectively learn long-range dependencies.

\subsection{Graph rewiring}

\textit{Graph Rewiring} is an effective method for addressing long-range interactions in graphs. 
It modifies connections between nodes by adding or adjusting edges and nodes, thus optimizing the network structure to enhance information propagation. 
Its capacity to capture global information has led to successful applications in various graph-level tasks such as Graph Classification (\cite{gilmer2017neural, pham2017graph, li2017learning, liu2022boosting}). 
In practice, graph rewiring is often combined with virtual nodes to enhance node connectivity and improve global information capture (\cite{li2017learning, hwang2021revisiting, qian2024probabilistic}). 
\citet{pham2017graph} introduced the Virtual Column Network incorporating a single fully-connected virtual node to learn representations of the entire graph. 
\citet{bruel2022rewiring} utilized positional encodings to rewire graphs with k-hop neighbors and virtual nodes. 
Recently, \citet{qian2023probabilistically} developed a probabilistic rewiring approach, sampling edges from a trainable distribution to strategically add relevant edges while excluding less beneficial ones. 
Subsequently, \citet{qian2024probabilistic} refined their method by incorporating virtual nodes for implicit rewiring, enhancing the transfer of long-range information.

Applying graph rewiring to time-series data, such as traffic prediction, presents unique challenges due to the dynamic nature of temporal dependencies. 
AGCRN (\cite{bai2020adaptive}) replaces the traditional distance-based adjacency matrix with an adaptive adjacency matrix tailored for time-series prediction, enabling the model to dynamically learn connectivity patterns based on temporal data. 
TimeGNN (\cite{xu2023timegnn}) utilizes Gumbel-Softmax to learn discrete adjacency matrices that represent the evolving graph structure at each time step, effectively capturing temporal changes without relying on a fixed structure. 
Similarly, \citet{chen2022balanced} employ a balanced graph structure learning method, which involves an adaptive adjacency matrix construction to balance the trade-off between local and global information capture, However, these methods neglect the information in the original adjacency matrix, potentially missing crucial static relationships. 
DGCRN (\cite{li2023dynamic}) integrates the original adjacency matrix into the adaptive one, leveraging both static and dynamic information, but its complexity and requirement for additional data, such as traffic speed, limit its practicality. 

This study proposes a method that incorporates virtual nodes to enhance connectivity and combines the benefits of both the original and adaptive adjacency matrices, improving the model's capacity to capture local and global temporal dependencies without additional data complexities.

\section{Methodology}
\label{sec:method}

\begin{figure}[htbp]
    \centering
    \includegraphics[width=\linewidth]{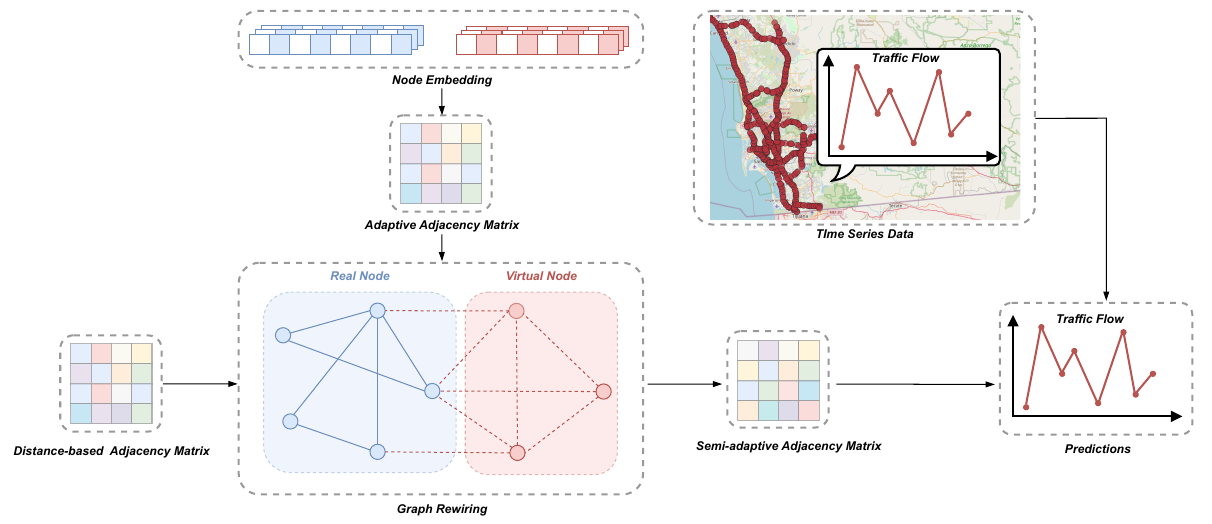}
    \caption{Traffic Prediction Framework Incorporating Virtual Nodes}
    \label{fig:flow_chart}
\end{figure}

In this section, we describe how to enhance the original graph with virtual nodes and apply this augmented graph for traffic prediction. 
The first part introduces the structure of STGCN (\cite{yu2017spatio}), which are selected as our predictive model $f_{\theta}$. 
Following this, we explain the incorporation of virtual nodes into the graph. We propose two approaches for integrating virtual nodes: using an adaptive adjacency matrix and a semi-adaptive adjacency matrix. 
The adaptive adjacency matrix is entirely task-driven, learning the optimal connections based on the prediction task, whereas the semi-adaptive adjacency matrix leverages both task-driven learning and geographical information to form connections. 
Combining two parts, the overall process of our method is illustrated in Figure \ref{fig:flow_chart}.

\subsection{Framework of STGCN}

Spatio-temporal graph convolutional network (STGCN) is designed to capture both spatial and temporal dependencies in traffic data (\cite{yu2017spatio}). 
The framework integrates Graph Convolutional Networks (GCNs) for spatial feature extraction and Temporal Convolutional Networks (TCNs) for temporal feature extraction. 
This combination allows STGCNs to effectively model the complex spatio-temporal dynamics inherent in traffic data.

Recall the notation in Section \ref{sec:problem}, let \(\mathcal{G} = (\mathcal{V}, \mathcal{E}, \mathbf{A})\) be a graph where \(\mathcal{V}\) is the set of nodes, \(\mathcal{E}\) is the set of edges, and \(\mathbf{A} \in \mathbb{R}^{|\mathcal{V}| \times |\mathcal{V}|}\) is the adjacency matrix. 
Given the node features matrix \(\mathbf{X} \in \mathbb{R}^{|\mathcal{V}| \times F}\), where \(F\) is the number of features, the spatial convolution operation in a GCN layer can be defined as:
\begin{equation}
\mathbf{H}^{(l+1)} = \sigma \left( \mathbf{A} \mathbf{H}^{(l)} \mathbf{W}^{(l)} \right),
\label{eq:gcn}
\end{equation}
where \(\mathbf{H}^{(l)}\) is the input to the \(l\)-th layer, \(\mathbf{W}^{(l)}\) is the weight matrix for the \(l\)-th layer, and \(\sigma\) is an activation function. 
The output of the GCN layer is then passed to the TCN layer to capture temporal dependencies.

The temporal convolution operation can be represented as:
\begin{equation}
\mathbf{Z}^{(t+1)} = \phi \left( \sum_{k=0}^{K-1} \mathbf{W}_k \mathbf{Z}^{(t-k)} \right),
\label{eq:tcn}
\end{equation}
where \(\mathbf{Z}^{(t)}\) is the input to the \(t\)-th time step, \(\mathbf{W}_k\) are the temporal convolutional filters, \(K\) is the kernel size, and \(\phi\) is the activation function. By stacking multiple GCN and TCN layers, STGCNs can effectively model both spatial and temporal aspects of traffic data.

In our study, we use STGCN as the base model \( f_{\theta} \) for short-term traffic prediction. 
The STGCN model takes as input the spatio-temporal data and the graph structure and outputs the predicted traffic conditions.
We further build models upon it to demonstrate the adaptation for long-term traffic prediction.

\subsection{Semi-adaptive adjacency matrix with virtual nodes}
In this section, we present our approach to enhance ST-GNN with virtual nodes, utilizing a semi-adaptive adjacency matrix. 
This method aims to effectively capture both local and global dependencies by integrating task-specific learning with geographic information.

We introduce virtual nodes to connect all existing real nodes $\mathcal{V}$ in the graph $\mathcal{G}$. 
The introduced virtual node set is denoted as $\mathcal{V}_{virtual}$, with $|\mathcal{V}_{virtual}|=n_v$, where \(n_v\) represents the number of virtual nodes. 
These virtual nodes are topologically connected to all existing real nodes, although the connection weights are initially unknown. 
This setup facilitates the aggregation of comprehensive information across the entire network, enhancing the model's ability to capture both local and global dependencies.

We then design an innovative adaptive learning mechanism that adjusts the connection weights between virtual nodes and real nodes based on the specific prediction task. 
Two node embeddings are assigned to each node, regardless of virtual or real nodes, resulting in two learnable embedding matrices \(\mathbf{E}_1\) and \(\mathbf{E}_2\) of dimensions \(\mathbb{R}^{(|\mathcal{V}| + n_v) \times d}\), where \(d\) is the embedding dimension, a hyperparameter.

\textbf{Adaptive Adjacency Matrix Construction:} We compute an anti-symmetric matrix \(\mathbf{A}_{adapt}\) as follows:
\begin{equation}
\mathbf{A}_{adapt} = \text{ReLU}(\mathbf{E}_1 \cdot \mathbf{E}_2^T - \mathbf{E}_2 \cdot \mathbf{E}_1^T).
\end{equation}
According to previous research (\cite{wu2020connecting}), the learned relation in time series forecasting is supposed to be uni-directional. 
The resulting \(\mathbf{A}_{adapt}\) is uni-directional. 
To remove weak connections, we apply a threshold \(r\), retaining only the elements in \(\mathbf{A}_{adapt}\) that exceed this threshold and setting the rest to zero. 
Mathematically, this can be represented as:
\begin{equation}
\mathbf{A}_{adapt, ij} = \begin{cases}
\mathbf{A}_{adapt, ij} & \text{if } \mathbf{A}_{adapt, ij} \geq r \\
0 & \text{otherwise},
\end{cases}
\end{equation}
where \(\mathbf{A}_{adapt, ij}\) represents the element in \(\mathbf{A}_{adapt}\) at position \((i, j)\).

\textbf{Semi-adaptive Adjacency Matrix Construction:} Given a distance-based adjacency matrix \(\mathbf{A}_{dist}\), we augment it with the virtual node connections from \(\mathbf{A}_{adapt}\). The semi-adaptive adjacency matrix \(\mathbf{A}_{semi}\) is constructed by integrating the adaptive matrix with the distance-based matrix:

\begin{equation}
\mathbf{A}_{semi} = \begin{bmatrix}
\mathbf{A}_{dist} & \mathbf{A}_{adapt, real\_to\_virtual} \\
\mathbf{A}_{adapt, virtual\_to\_real} & \mathbf{A}_{adapt, virtual\_nodes}
\end{bmatrix}.
\end{equation}

The relationships between \(\mathbf{A}_{adapt, real\_to\_virtual}\), \(\mathbf{A}_{adapt, virtual\_to\_real}\), \(\mathbf{A}_{adapt, virtual\_nodes}\), and \(\mathbf{A}_{adapt}\) are given by:
\begin{equation}
\left\{
\begin{aligned}
    &\mathbf{A}_{adapt, real\_to\_virtual} = \mathbf{A}_{adapt}[1:|\mathcal{V}|, |\mathcal{V}|+1:|\mathcal{V}|+n_v] \\
    &\mathbf{A}_{adapt, virtual\_to\_real} = \mathbf{A}_{adapt}[|\mathcal{V}|+1:|\mathcal{V}|+n_v, 1:|\mathcal{V}|] \\
    &\mathbf{A}_{adapt, virtual\_nodes}    = \mathbf{A}_{adapt}[|\mathcal{V}|+1:|\mathcal{V}|+n_v, |\mathcal{V}|+1:|\mathcal{V}|+n_v]
\end{aligned}
\right.
\end{equation}

In this matrix, \(\mathbf{A}_{adapt, real\_to\_virtual}\) and \(\mathbf{A}_{adapt, virtual\_to\_real}\) are the parts of \(\mathbf{A}_{adapt}\) that correspond to the connection weights between real nodes and virtual nodes. 
This approach ensures that the semi-adaptive adjacency matrix leverages both the inherent geographical information and the task-specific learned relationships.

\textbf{Virtual Node Signals Initialization:} To ensure computational simplicity, we initialize the virtual node signals to zero, indicating that virtual nodes start with no traffic flow information at the initial time step. 
This allows the virtual nodes to gradually learn and aggregate relevant information from the real nodes through the adaptive learning process.

\section{Experiment}
\label{sec:experiment}
\subsection{Dataset}
\label{subsec:dataset}
The dataset used for this research is the San Diego (SD) sub-dataset sourced from the LargeST benchmark dataset (\cite{liu2023largest}). 
This dataset provides a comprehensive collection of traffic data from 716 sensors located in San Diego County, California. These sensors cover a variety of highways and collect data on traffic flow. The data spans from January 1, 2017, to December 31, 2021, providing a rich temporal coverage to study traffic patterns and their fluctuations.

In LargeST dataset, the adjacency matrix $\mathbf{A}$ is given through a thresholded Gaussian kernel (\cite{shuman2013emerging}), where $\mathbf{A}_{ij} = \exp(-\frac{d_{ij}^2}{\sigma^2})$ if $\mathbf{A}_{ij} \geq r$, otherwise $\mathbf{A}_{ij} = 0$. 
Here, $d_{ij}$ denotes the road network distance between nodes $i$ and $j$, $\sigma$ is the standard deviation of all distances, and $r$ is the threshold. 
Consequently, the adjacency matrix $\mathbf{A}$ reflects the geographical affinities of the regions, with shorter distances indicating larger adjacency values.

\begin{figure}[htbp]
    \centering
    \begin{minipage}{0.5\textwidth}
        \centering              
        \begin{tabular}{lc}
            \toprule
            \textbf{Attribute} & \textbf{Value} \\
            \midrule
            Nodes & 716 \\
            Edges & 17,319 \\
            Average Degree & 24.2 \\
            Density & 0.0338 \\
            Time Range & 01/01/2017 – 12/31/2021 \\
            Sampling Rate & 5 minutes \\
            Time Frames & 525,888 \\
            Data Points & 0.38B \\
            \bottomrule
        \end{tabular}  
        \captionof{table}{Detailed Characteristics of the SD Dataset. B: billion ($10^9$)}
        \label{tab:dataset_summary}
    \end{minipage}
    \hfill
    \begin{minipage}{0.4\textwidth}
        \centering
        \includegraphics[width=\textwidth]{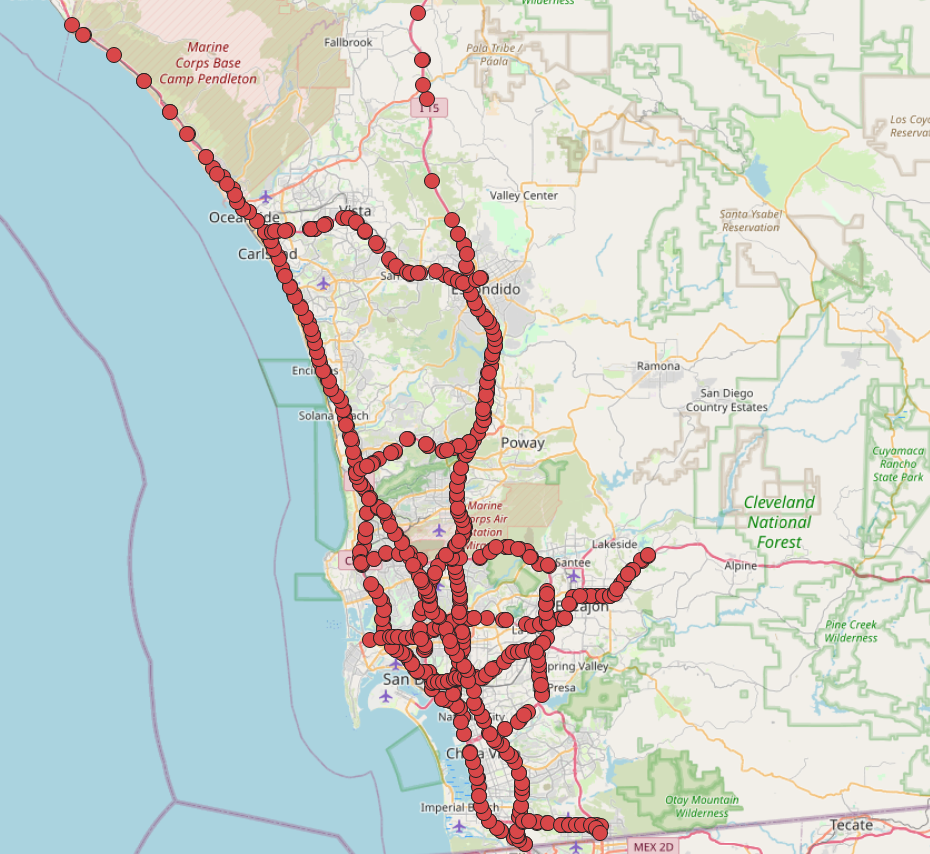}
        \caption{Visualization of Sensor Locations in San Diego}
        \label{fig:sensor_map}
    \end{minipage}
\end{figure}

The SD sub-dataset, being the smallest in the LargeST collection, was specifically chosen to strike an optimal balance between computational demand and the complexity of model parameters. 
This choice addresses potential overfitting issues that could arise when virtual nodes are added to small-sized graphs where information propagation is already efficient without additional enhancements. 
The SD dataset, with its 716 nodes and 17,319 edges, offers a moderately scaled environment ideal for testing the impact of virtual nodes. 
Its detailed descriptions are listed in Table \ref{tab:dataset_summary} and the spatial distributions of all sensors are given in Figure \ref{fig:sensor_map}.

\begin{figure}[htbp]
    \centering
    \includegraphics[width=0.5\linewidth]{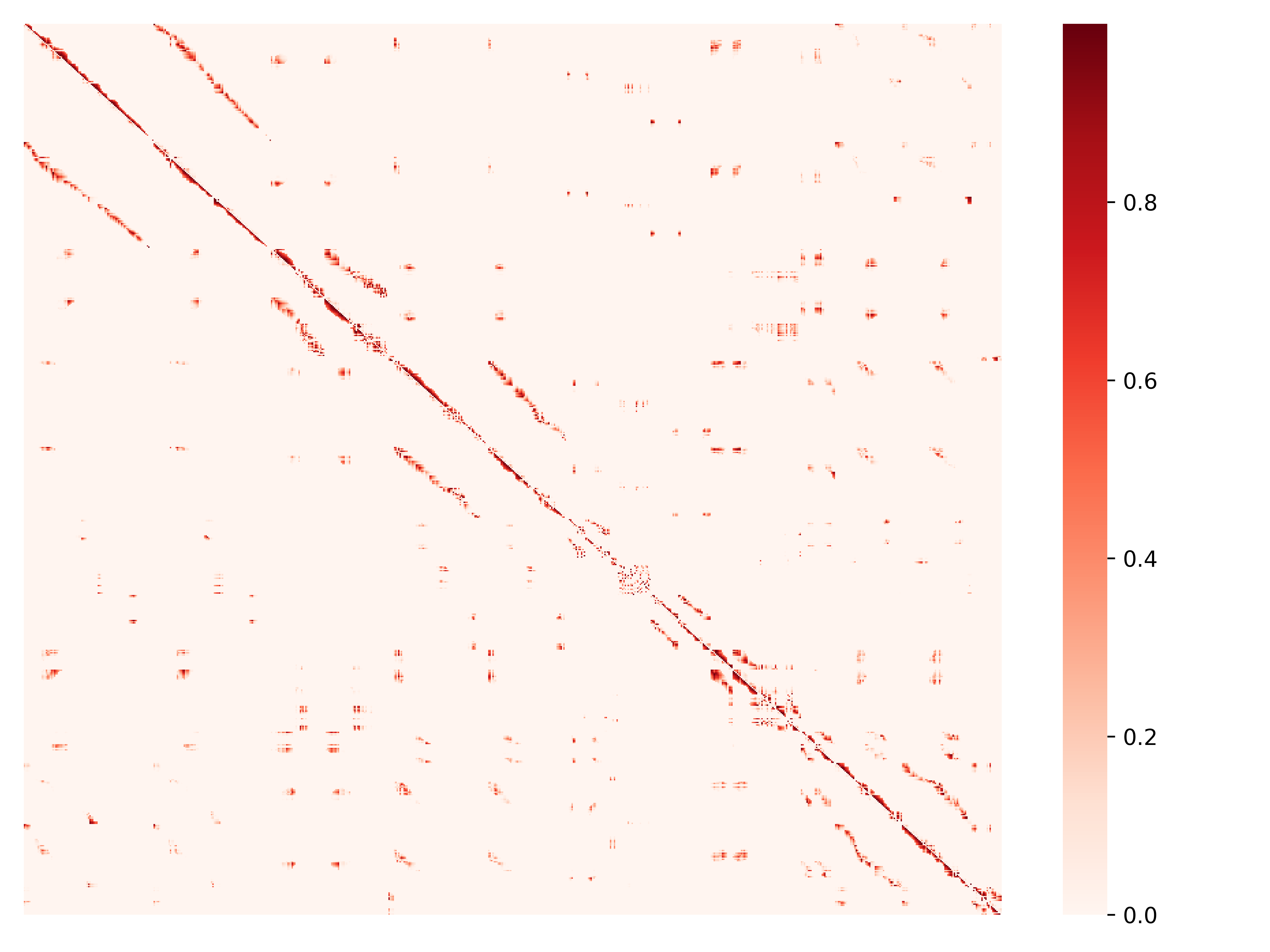}
    \caption{Adjacency Matrix Heat Map}
    \label{fig:sd_adj}
\end{figure}

 Despite the relatively large number of nodes in the SD dataset, its graph structure remains sparse, with a graph density of only 0.03, which inhibits effective information transfer across the graph. 
 This sparsity is also evident from the adjacency matrix.
 Figure \ref{fig:sd_adj} presents the adjacency matrix heatmap, which highlights the sparse connectivity of the graph; direct links are absent for most nodes, with only isolated clusters exhibiting stronger connections. 
 The sparsity is a trade-off in GNN training, which accelerates the GNN training while limiting information propagation, necessitating multiple intermediate nodes for transmission across the graph. 
 The integration of virtual nodes mitigates this limitation by establishing direct pathways, thereby enhancing information flow efficiency.

\subsection{Baseline models and evaluation metrics}
\label{sec:baseline}
In our study, we aim to assess the impact of various adjacency matrix configurations on the performance of our predictive model. Naturally, the original STGCN is one of the baseline models.

Moreover, We employ a modified "All-ones" setting as a baseline where only the rows and columns corresponding to a single virtual node in the adjacency matrix are set to ones, while the remaining parts are still distance-based. 
Specifically, that means we assign $\mathbf{A}_{adapt, real\_to\_virtual} = \mathbf{1}$, $\mathbf{A}_{adapt, virtual\_to\_real} = \mathbf{1}$.
This approach simplifies the interaction framework to some extent while preserving the essential distance-based connections. 
This baseline is contrasted with adaptive and semi-adaptive matrices that are specifically designed to capture the dynamic interconnections and contextual relationships among data points.

To evaluate the performance of these configurations, we use two primary metrics: Root Mean Squared Error (RMSE) and Mean Absolute Percentage Error (MAPE). 
These metrics are widely used in the literature for their effectiveness in measuring prediction accuracy and error.

The RMSE is defined as:
\begin{equation}
\text{RMSE} = \sqrt{\frac{1}{n} \sum_{i=1}^{n} (y_i - \hat{y}_i)^2},
\end{equation}
where \(y_i\) represents the observed values, \(\hat{y}_i\) represents the predicted values, and \(n\) is the number of observations. RMSE provides a measure of the average magnitude of the prediction errors, giving higher weight to larger errors.

The MAPE is defined as:
\begin{equation}
\text{MAPE} = \frac{1}{n} \sum_{i=1}^{n} \left| \frac{y_i - \hat{y}_i}{y_i} \right|.
\end{equation}
MAPE expresses the prediction accuracy as a percentage, making it easier to interpret the model's performance relative to the actual values.

\subsection{Model comparison}
\label{sec:model_comparison}
We apply different adjacency matrix configurations on STGCN (\cite{yu2017spatio}) to test whether virtual nodes help long-term traffic prediction. 
Notably, our approach is applicable to any ST-GNN or other graph models accepting adjacency matrices as input, we will include more models in our future studies.

We choose various adjacency matrix configurations for comparison:
\begin{itemize}
    \item A distance-based pre-defined adjacency matrix serves as the baseline.
    \item The "All-ones" case introduced in Section \ref{sec:baseline}.
    \item An adaptive adjacency matrix with virtual nodes derived from node embeddings.
    \item A semi-adaptive adjacency matrix that integrates both pre-defined and adaptive matrices.
\end{itemize}

For the latter two configurations, we assessed the effects of introducing between one and twenty virtual nodes on prediction accuracy. 
We utilized time series data from the years 2019 to 2020 for training and testing, encompassing a total of 35,040 time frames. 
We evaluated the model's predictive performance across 1 to 20 prediction horizons, where each horizon corresponds to a five-minute interval, the same as the sampling rate.

All our experiments are implemented on a machine with Ubuntu 22.04, with Intel(R) Core(TM) i9-10980XE CPU @ 3.00GHz CPU, 128GB RAM, and NVIDIA GeForce RTX 4080 GPU.

Table \ref{tab:performance} presents the performance metrics of various adjacency matrix configurations over multiple prediction horizons (Horizon 5, Horizon 10, Horizon 15, Horizon 20) and average prediction horizons (Average, Average (75-100 min.)). 
We use V.N. as an abbreviation to denote different quantities of virtual nodes in various models.
Across all horizons and the average metrics, \textbf{Semi-10 V.N.} consistently provides the lowest RMSE and MAPE, making it the most effective configuration for enhancing prediction accuracy. 
This hybrid approach, leveraging both static and dynamic adjacency matrix components, appears to strike the optimal balance for the STGCN model.

\begin{table}[htbp]
    \scriptsize
    \setlength{\tabcolsep}{4.5pt} 
    \centering
    \begin{tabular}{l|cc|cc|cc|cc|cc|cc}
        \toprule
        \multirow{2}{*}{Adjacency Matrix} & \multicolumn{2}{c|}{Horizon 5} & \multicolumn{2}{c|}{Horizon 10} & \multicolumn{2}{c|}{Horizon 15} & \multicolumn{2}{c|}{Horizon 20} & \multicolumn{2}{c|}{Average} & \multicolumn{2}{c}{Average (75-100 min.)} \\
        \cmidrule(r){2-13}
         & RMSE & MAPE & RMSE & MAPE & RMSE & MAPE & RMSE & MAPE & RMSE & MAPE & RMSE & MAPE \\
        \midrule
        Distance-based   & 35.35 & 0.1383 & 39.63 & 0.1566 & 44.10 & 0.1791 & 45.89 & 0.1839 & 39.73 & 0.1590 & 45.15 & 0.1827 \\
        All-ones         & 34.76 & 0.1348 & 39.24 & 0.1566 & 43.32 & 0.1727 & 46.55 & 0.1825 & 39.29 & 0.1554 & 45.05 & 0.1785 \\
        Adaptive-1 V.N.  & 36.20 & 0.1487 & 41.39 & 0.1664 & 46.09 & 0.1797 & 49.13 & 0.1876 & 41.31 & 0.1647 & 47.74 & 0.1841 \\
        Adaptive-2 V.N.  & 35.74 & 0.1473 & 41.30 & 0.1665 & 46.15 & 0.1795 & 49.17 & 0.1877 & 41.12 & 0.1645 & 47.68 & 0.1835 \\
        Adaptive-5 V.N.  & 35.68 & 0.1451 & 41.63 & 0.1658 & 46.42 & 0.1788 & 48.62 & 0.1838 & 41.31 & 0.1632 & 47.76 & 0.1821 \\
        Adaptive-10 V.N. & 35.72 & 0.1453 & 41.16 & 0.1662 & 45.88 & 0.1807 & 49.05 & 0.1874 & 41.00 & 0.1639 & 47.54 & 0.1842 \\       
        Adaptive-20 V.N. & 35.97 & 0.1461 & 41.38 & 0.1640 & 46.02 & 0.1782 & 48.66 & 0.1834 & 41.18 & 0.1625 & 47.47 & 0.1811 \\
        Semi-1 V.N.      & 35.26 & 0.1436 & 39.92 & 0.1635 & 44.25 & 0.1737 & 47.26 & 0.1804 & 40.03 & 0.1590 & 45.88 & 0.1775 \\
        Semi-2 V.N.      & 35.36 & 0.1414 & 39.88 & 0.1657 & 43.49 & 0.1741 & 45.56 & 0.1812 & 39.56 & 0.1599 & 44.61 & 0.1776 \\
        Semi-5 V.N.      & 36.20 & 0.1522 & 41.80 & 0.1789 & 46.62 & 0.1891 & 49.16 & 0.1926 & 41.53 & 0.1717 & 47.95 & 0.1921 \\
        Semi-10 V.N.     & \textbf{34.10} & \textbf{0.1346} & \textbf{38.06} & \textbf{0.1558} & \textbf{41.28} & \textbf{0.1709} & \textbf{43.38} & \textbf{0.1750} & \textbf{37.82} & \textbf{0.1537} & \textbf{42.32} & \textbf{0.1735} \\
        Semi-20 V.N.     & 36.04 & 0.1460 & 40.79 & 0.1639 & 45.55 & 0.1851 & 47.91 & 0.1890 & 40.82 & 0.1644 & 46.89 & 0.1870 \\
        \bottomrule
    \end{tabular}
    \caption{Performance of Different Adjacency Matrix Configurations}
    \label{tab:performance}
\end{table}

\begin{figure}[htbp]
    \centering
    \includegraphics[width=\linewidth]{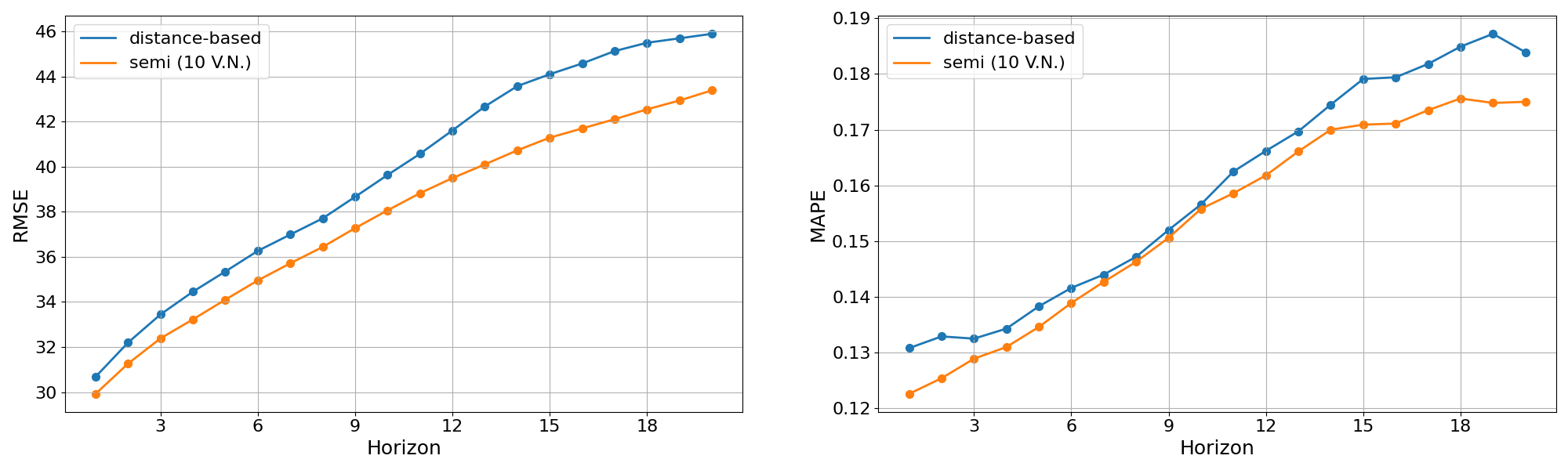}
    \caption{Long-term Performance of Virtual Nodes}
    \label{fig:long_terms}
\end{figure}

The "All-ones" configuration shows a marginal improvement in both the average and long-term (75-100 min.) results, suggesting that employing a uniform structure, which includes virtual nodes, can be potentially beneficial for traffic prediction. 
However, at specific horizons, the "All-ones" configuration does not perform as well as the "Distance-based" configuration, indicating that there is still room for optimization in these settings.

The introduction of virtual nodes significantly enhances long-term prediction accuracy. 
Specifically, the \textbf{Semi-10 V.N.} configuration achieves an RMSE of 42.32 and a MAPE of 0.1735 in the 75-100 minute average horizon, representing an RMSE reduction of approximately 6.27\% and a MAPE reduction of approximately 5.04\% compared to the distance-based baseline.

Figure \ref{fig:long_terms} illustrates the long-term performance of virtual nodes by comparing the RMSE and MAPE over varying prediction horizons for the distance-based and semi-adaptive (10 V.N.) configurations. 
The line plots indicate that the semi-adaptive configuration consistently outperforms the distance-based configuration across all horizons. 
The semi-adaptive (10 V.N.) configuration consistently shows lower RMSE and MAPE among all horizons. 
The gaps between the two configurations in both RMSE and MAPE widen as the prediction horizon increases. 
Overall, the introduction of virtual nodes enhances both short-term and long-term predictions, with particularly noticeable improvements in long-term performance.

\subsection{Sensitivity analysis}
\begin{figure}[htbp]
    \centering
    \begin{subfigure}{\linewidth}
        \centering
        \includegraphics[width=\linewidth]{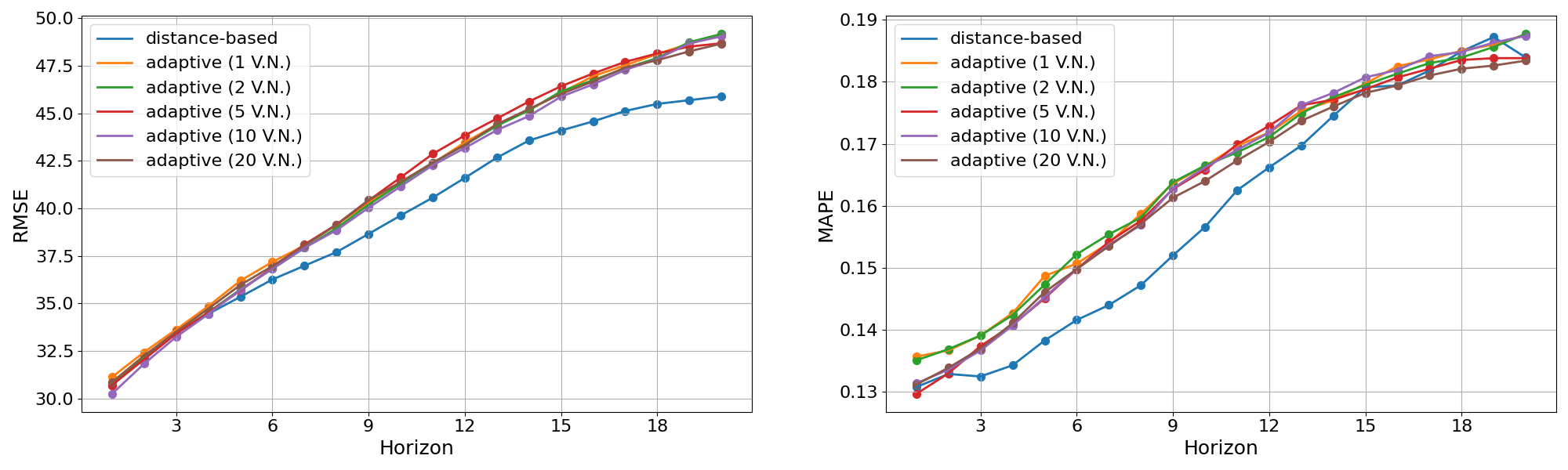}
        \caption{Adaptive Adjacency Matrix}
        \label{fig:sensitivity-adaptive}
    \end{subfigure}
    \begin{subfigure}{\linewidth}
        \centering
        \includegraphics[width=\linewidth]{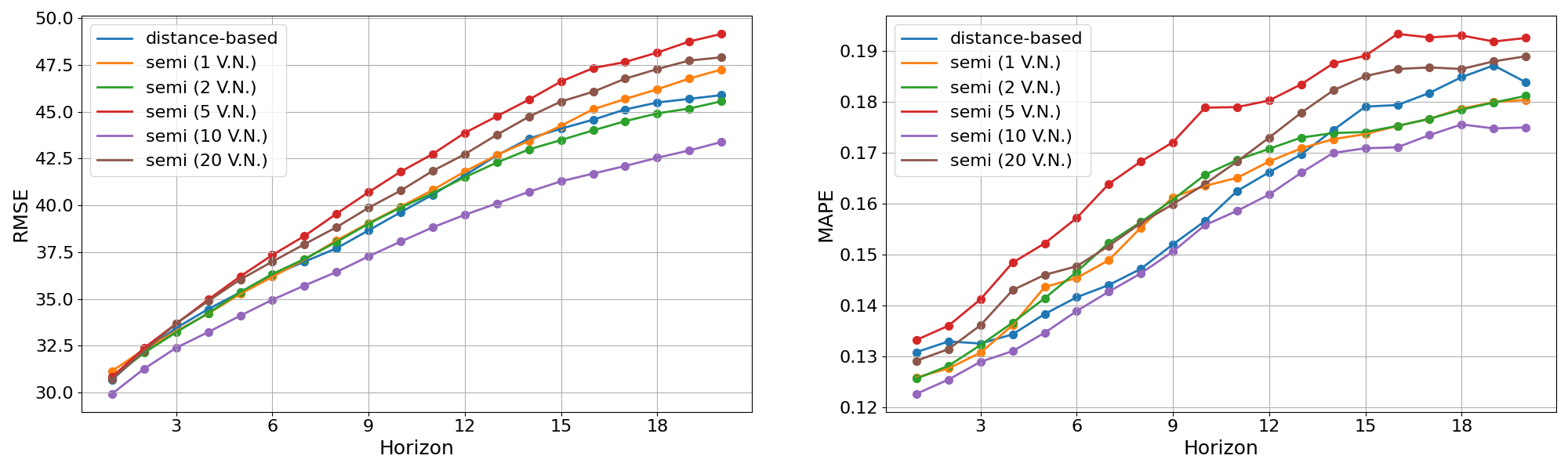}
        \caption{Semi-adaptive Adjacency Matrix}
        \label{fig:sensitivity-semi}
    \end{subfigure}
    \caption{Sensitivity Analysis of Virtual Nodes Number in Adaptive and Semi-Adaptive Adjacency Matrices}
    \label{fig:sensitivity-analysis}
\end{figure}

The sensitivity analysis focuses on evaluating the impact of varying the number of virtual nodes on the performance of both adaptive and semi-adaptive configurations. 
Figures \ref{fig:sensitivity-analysis} provide insights into this sensitivity analysis, illustrating the RMSE and MAPE metrics for different numbers of virtual nodes across multiple prediction horizons.

\textbf{Adaptive Adjacency Matrix:} Figure \ref{fig:sensitivity-adaptive} displays the RMSE and MAPE trends for adaptive configurations with 1, 2, 5, 10, and 20 virtual nodes compared to the distance-based baseline. 
The distance-based configuration exhibits the lowest RMSE and MAPE across all prediction horizons, suggesting that the adaptive configurations, despite their complexity, do not outperform the simpler distance-based approach. 
As the number of virtual nodes increases, there is no significant improvement in either RMSE or MAPE, indicating that the additional parameters introduced by virtual nodes do not translate to better performance. 
In fact, the adaptive configurations show higher error rates, which may imply overfitting and instability due to the increased model complexity. Therefore, we need to limit the complexity of the adjacency matrix, prompting us to explore semi-adaptive configurations.

\textbf{Semi-adaptive Adjacency Matrix:} Figure \ref{fig:sensitivity-semi} presents the RMSE and MAPE trends for semi-adaptive configurations. The semi-adaptive configurations generally show better performance than both the adaptive configurations and the distance-based baseline across all prediction horizons. As the number of virtual nodes increases from 1 to 10, the overall performance improves, with the configurations of 1 and 2 virtual nodes already surpassing the baseline in some horizons. Notably, the Semi-10 V.N. configuration demonstrates significantly lower RMSE and MAPE compared to other semi-adaptive settings and the baseline, particularly in longer horizons. This indicates that a balanced complexity, as seen with the Semi-10 V.N. configuration, provides the optimal enhancement in prediction accuracy. When the number of virtual nodes reaches 20, the performance declines, suggesting that the optimal number of virtual nodes is around 10.

\subsection{Visualization of virtual nodes}

In this section, we visualize the virtual nodes' connection weights and spatial patterns to explore their relationship with real-world transportation networks. 
The visualization provides insights into how virtual nodes integrate with actual road networks and help in understanding their impact on transportation modeling and prediction accuracy.

\begin{figure}[htbp]
    \centering
    \begin{subfigure}[b]{0.55\linewidth}
        \centering
        \includegraphics[width=\linewidth]{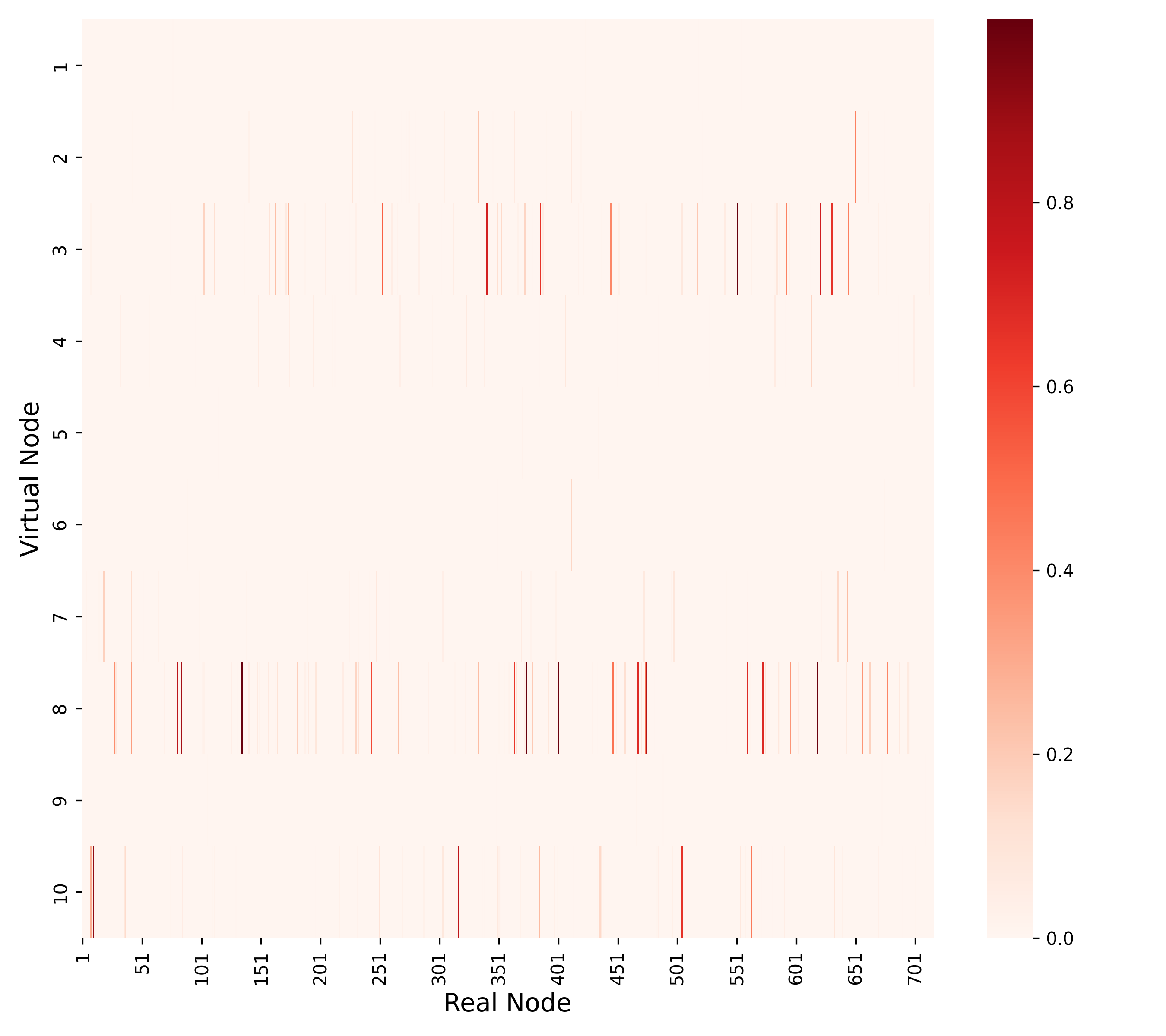}
        \caption{Real-to-virtual Adjacency Matrix Heat Map}
        \label{fig:r2v}
    \end{subfigure}
    \hfill
    \begin{subfigure}[b]{0.4\linewidth}
        \centering
        \includegraphics[width=\linewidth]{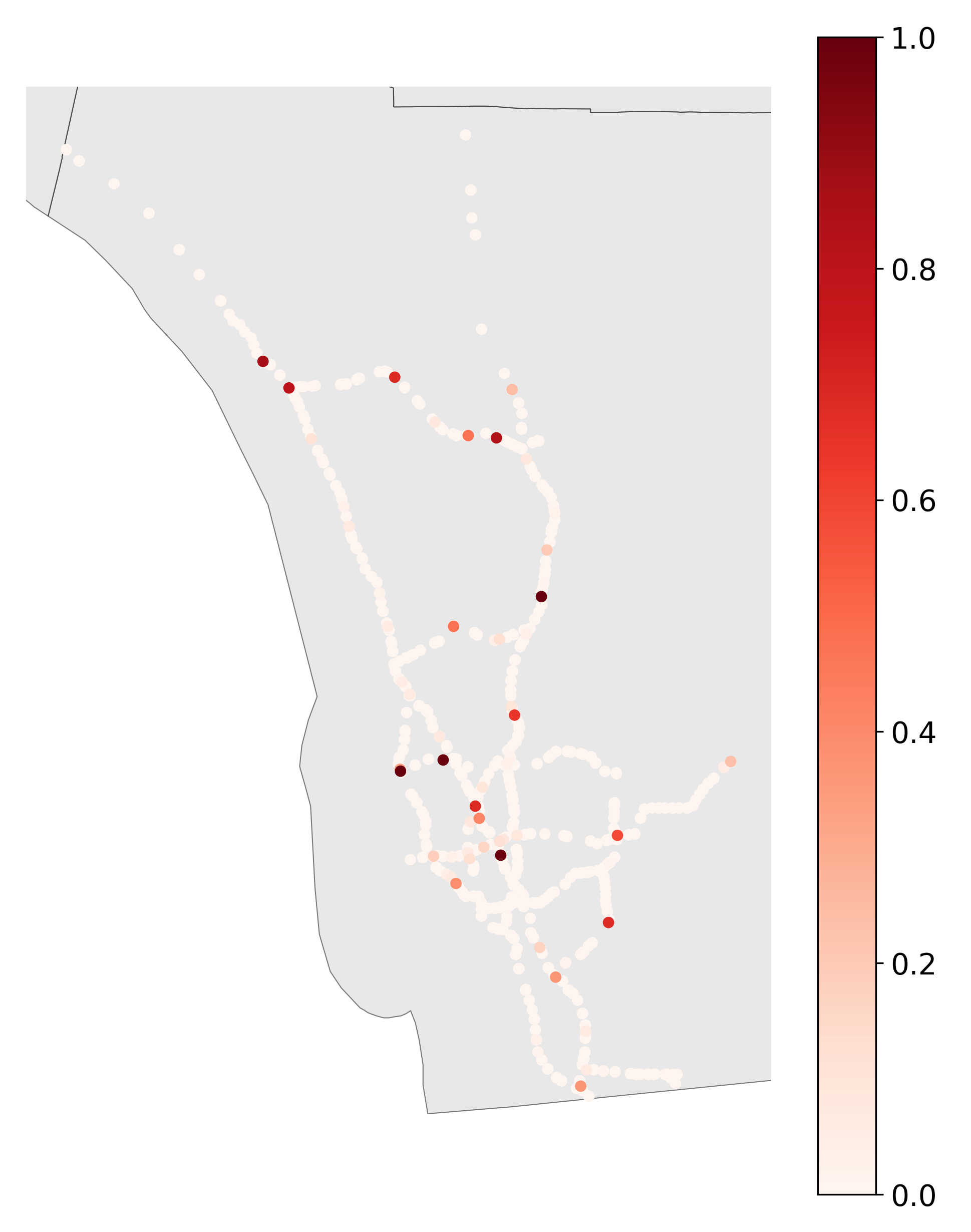}
        \caption{Road Network Heat Map of Virtual Node 8}
        \label{fig:vn8}
    \end{subfigure}
    
    \begin{subfigure}[b]{\linewidth}
        \centering
        \includegraphics[width=\linewidth]{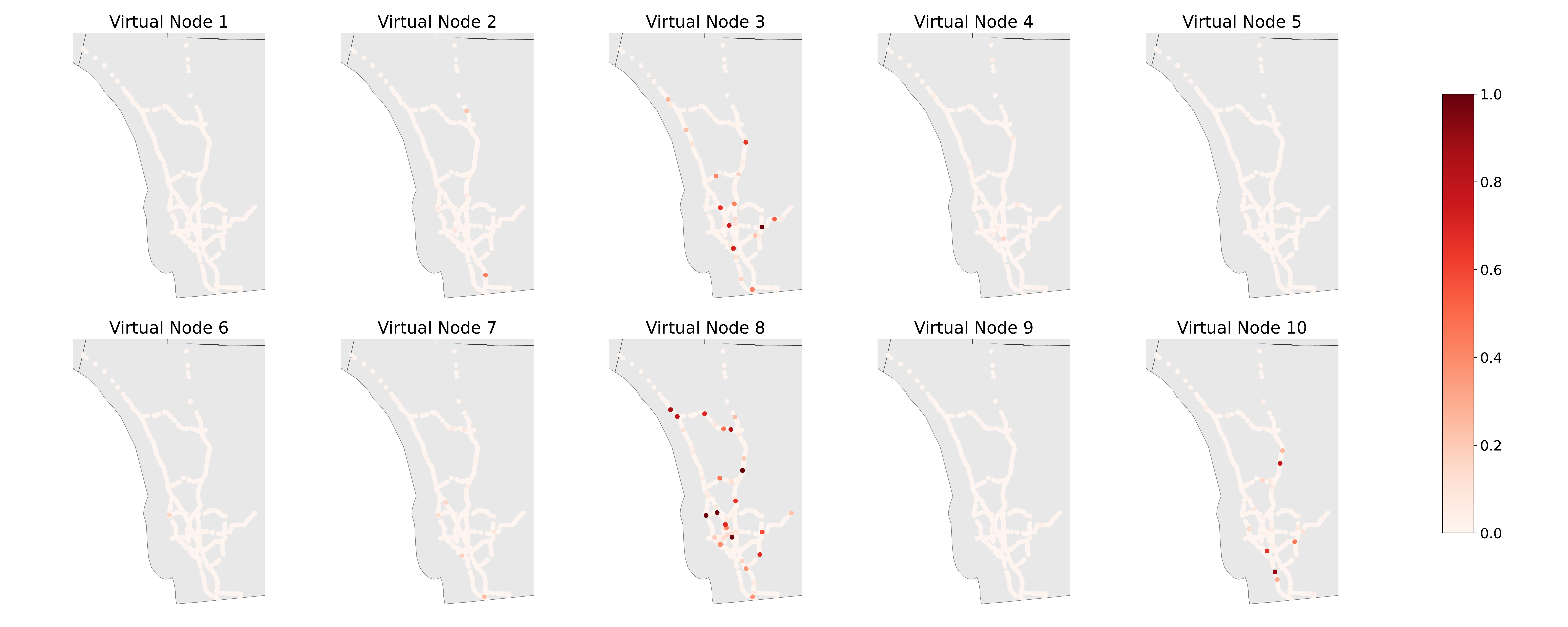}
        \caption{Road Network Heat Map of All Virtual Nodes}
        \label{fig:vnall}
    \end{subfigure}
    \caption{Visualization of virtual node connection weights.}
    \label{fig:combined}
\end{figure}

\textbf{Real-to-Virtual Adjacency Matrix:} We begin by visualizing the adjacency matrix that captures the connections between real nodes and virtual nodes. 
Figure \ref{fig:r2v} shows the heat map of this adjacency matrix for the semi-adaptive configuration with ten virtual nodes, which corresponds to the model with the best prediction performance. 
Each row in the heat map corresponds to a virtual node, and each column corresponds to a real node. 
The intensity of the color represents the strength of the connection between the real and virtual nodes. 
Darker colors indicate stronger connections. 
The heat map reveals that certain virtual nodes have stronger connections with specific real nodes, indicated by the darker bands in the matrix. 
Notably, Virtual Nodes 3, 8, and 10 show significant connectivity with a wide range of real nodes, with Virtual Node 8 demonstrating the strongest connections. 
This suggests that these three virtual nodes are more active and play a crucial role in aggregating information across the entire network.

\textbf{Road Network Heat Map:} Next, we visualize the road network by mapping the connection strengths from the adjacency matrix onto the network. By mapping the real-to-virtual adjacency matrix onto a road network, we can identify key areas where virtual nodes influence traffic patterns. Each subfigure in Figure \ref{fig:vnall} represents a virtual node, and the color intensity on the road network shows the connection strength. The visualization results reinforce the previous findings: Virtual Nodes 3, 8, and 10 have stronger connections with multiple regions of the road network, while the remaining virtual nodes exhibit relatively weaker connections, suggesting that Virtual Nodes 3, 8, and 10 play crucial roles in aggregating information from the real nodes, thereby enhancing the overall prediction accuracy of the model.

\textbf{Spatial Patterns of Connection Weights:} To delve deeper into the spatial distribution of connection weights, we examine the patterns of these weights across the road network. Understanding these spatial patterns can help explain why virtual nodes improve prediction accuracy. Taking Virtual Node 8 as an example, we observe that the real nodes with higher connection weights are predominantly located at intersections and key junctions in the road network. Figure \ref{fig:vn8} illustrates this pattern clearly. This distribution indicates that Virtual Node 8 has effectively learned to assign greater importance to areas with higher traffic activity. These intersections are critical points in the network where traffic flows converge and diverge, making them vital for accurate traffic prediction. By weighting these busy areas more heavily, Virtual Node 8 can better capture the dynamics of traffic flow, leading to more precise and reliable predictions.

This spatial pattern is not unique to Virtual Node 8. Similar trends are observed for Virtual Nodes 3 and 10, which also show higher weights concentrated around key intersections and junctions. This automatic learning process demonstrates the virtual nodes' ability to identify and prioritize essential areas within the road network, thereby enhancing the model's overall effectiveness.

\section{Discussion and conclusions}
\label{sec:conclusion}
In this study, we introduced a novel approach to enhance long-term traffic prediction by incorporating virtual nodes into existing ST-GNN models. 
The introduction of virtual nodes addresses the \textit{over-squashing} problem inherent in traditional GNNs, which limits their ability to capture long-range dependencies. 
By connecting virtual nodes to all real nodes, we enable more efficient message-passing process across the graph, enhancing the model's capability to capture critical spatial relationships and long-term correlations. 
The semi-adaptive adjacency matrix, which combines task-specific learning with geographical information, helps the learning of virtual nodes' connection weights and provides explainable results to decision-makers.

Our findings show that integrating virtual nodes, particularly with the semi-adaptive adjacency matrix, significantly improves long-term traffic prediction accuracy. 
Virtual nodes enhance information aggregation and optimize information flow through dynamic weight adjustment.
Additionally, the visualization of the adjacency matrix and road network heat maps shows how virtual nodes prioritize key intersections and traffic-active areas, enhancing explainability.

Despite these advancements, there are several directions for future research. 
First, incorporating time-varying adjacency matrices to better capture dynamic relationships in spatio-temporal data is a promising direction. 
Second, testing the scalability of our approach on larger datasets and more complex networks is essential to validate its practicality in real-world applications. 

In conclusion, the incorporation of virtual nodes represents a significant step forward in enhancing long-term traffic prediction. 
Our approach not only addresses the limitations of traditional ST-GNNs but also provides a flexible and robust solution for capturing complex spatio-temporal dynamics. 
This advancement has the potential to significantly improve traffic management and urban planning, ultimately contributing to more efficient and sustainable transportation systems.

\bibliography{references}
\bibliographystyle{unsrtnat}

\end{document}